\def\pasubmission{0}

\if1\pasubmission
\documentclass[12pt]{article}
\else
\documentclass[11pt]{article}
\fi

\usepackage[letterpaper, left=1.25in, right=1.25in, top=1.2in, bottom=1.2in]{geometry}

\usepackage[colorlinks=true, citecolor=blue, linkcolor=blue, urlcolor=blue]{hyperref}
\usepackage[dvipsnames]{xcolor}

\usepackage{natbib} %bib
\usepackage{caption}
\usepackage{subcaption}

\usepackage{amsmath, amssymb, mathtools, bbm}
\usepackage{amsfonts}
\usepackage{bm} %write letters in bold in math mode
\usepackage{physics}  
\usepackage{mathpazo}

\usepackage{amsthm}

\theoremstyle{plain}

\theoremstyle{definition}

\theoremstyle{remark}

\usepackage{multirow} %for merging table rows
\usepackage{float}

\usepackage{tikz}
\usetikzlibrary{shapes, arrows.meta, positioning, calc}

\usepackage{graphicx}
\usepackage{booktabs}
\usepackage{microtype}
\usepackage{setspace}
\usepackage{placeins}
\usepackage{comment}

\makeatletter
\newcommand{\assumplabel}[2]{%
   \protected@write \@auxout {}{\string\newlabel{#1}{{#2}{\thepage}{#2}{#1}{}}}%
   \hypertarget{#1}{#2}%
}
\makeatother

\usepackage{longtable}
\if1\pasubmission
\fi

\usepackage{float}

\newcommand{\tit}{Gendered Misclassification from Surname-Based Race Imputation}
\title{{\bf\tit}}

\author{Noah Dasanaike\thanks{PhD Candidate, Department of Government, Harvard University. Corresponding author: \href{mailto:noahdasanaike@g.harvard.edu}{noahdasanaike@g.harvard.edu}.}}
\date{\today}

\begin{document}

\maketitle

\begin{abstract}
Researchers increasingly estimate political behavior based on race and gender from voter files, where race is imputed from surnames and geography. This only works where surnames suggest group membership, an assumption violated where surnames follow norms around marriage. Imputation is therefore biased against women who adopt their husband's surname, especially those who racially inter-marry, an outcome itself endogenous to partisanship. Using six US states recording self-reported race, I show that imputation is less accurate for Asian and Hispanic women, particularly those sharing the surname of a partner of another race, who differ politically from other women of their race. This violates assumptions behind downstream inference approaches, which nearly double the gender gap in partisanship among Asian voters. I derive an expression showing the bias is largest where a group is small. Using maiden rather than marriage-adopted surnames in North Carolina raises accuracy up to fourfold for women married across racial lines. Recovering maiden surnames, or omitting the married surname, removes much of the bias. In a national replication of a study of gender gaps in turnout, the correction more than triples the gap among Asian voters. In Lebanon, where wives register in their husband's locality, the error enters through geography. \\

\end{abstract}
\newpage

\section{Introduction}

Studies of political behavior and other individual-level quantities of interest, both in the United States and elsewhere, increasingly rely on records for which race is imputed. The most common approach for doing so, BISG, involves assigning each individual a probability of belonging to each racial or ethnic category based on their surname and location of residence \citep{elliott2009using, ImaiKhanna2016, voicu2018firstname, ImaiOlivellaRosenman2022sciadv, dasanaike2026embedding, chasalow2026lbisg}. These probabilities are then used to estimate turnout, partisanship, income, health outcomes, and other quantities by group, and increasingly by group and sex together \citep{AnsolabehereHersh2013, GhitzaGelman2020, StaufferFraga2022, GrumbachSahnStaszak2022}. Most recently, \citet{mccartan2025estimating} show that the outcome of interest itself can be used to correct the bias induced by these probabilities, so long as the outcome is unrelated to the surname once group membership and residence are known. 

However, this approach implicitly assumes that surname is suggestive of racial or ethnic group membership. Yet, surnames, and the records to which they are attached, often follow rules of marriage rather than ethno-racial inheritance alone. In the United States, for instance, most married women adopt the surname of their husband \citep{goldinshim2004making, pew2023namechange}, and intermarriage is itself disproportionately common among certain groups (such as Asian and Hispanic women) and among those of certain partisan attachments  \citep{pew2017intermarriage, QianLichter2011}. According to any approach that draws on surnames, the surname of a woman in group $x$ married to a  man in group $y$ is therefore, for the purposes of imputation, the surname of group $y$. More generally, wherever the rules or social norms that assign surnames differ by gender, probabilistic inferences of group membership based on surname inherit a bias that is concentrated among women. Moreover, in the case of marriage, where the bias depends on whom women marry, the question of whom people marry is itself related to their politics. The bias is thus correlated with downstream outcomes that researchers are interested in. 

This possibility has long been noted, though largely in passing. The Census Bureau, for instance, excluded ever-married women when constructing the 1990 Spanish surname list \citep{WordPerkins1996}, and cancer registries classify women by their maiden surname where it is recorded \citep{NAACCR2011nhia}. \citet{elliott2009using} and \citet{argylebarber2024misclassification} both note briefly that surname-based imputation is less accurate for Hispanic and Asian women than for Hispanic and Asian men, which they both attribute to surname change at marriage. \citet{ImaiKhanna2016} similarly note that inter-racial marriage rates that differ by gender could violate the assumptions of BISG. Yet, no study has systematically measured this bias, either writ-large or broken down by marital status and the race of the spouse, estimated the consequences for the race-based estimates that researchers obtain downstream, or proposed a correction for the problem. 

In this paper, I do so in the United States using the six states whose voter files record self-reported race (Alabama, Florida, Georgia, Louisiana, North Carolina, and South Carolina). In these states, I observe the race of more than 38 million registered voters, the race of their inferred co-resident partners, and, in three states, their party registration. First, I demonstrate a lower accuracy of surname-based imputation for women than for men, and demonstrate that the lower accuracy for Asian and Hispanic women is concentrated among women who share the surname of a partner of another race. Second, I demonstrate that these women differ politically from other women of their own racial group. Their misclassification thus violates the assumption that the outcome is unrelated to the surname given race. As a result, methods for estimating downstream quantities from racial probabilities, BIRDiE in particular, nearly double the gender gap in partisanship among Asian registered voters.

Third, I derive an expression for this bias as the product of how common married surnames of another race are, how much less likely imputation is to assign the women with those surnames to their own race, and how much these women differ from other women of their race. Combined with census data on intermarriage, the expression shows that the bias is largest where a group is small, which is also the setting where surveys are least able to substitute for voter files. Fourth, I show that recovering maiden surnames from middle names, which I validate using two decades of voter files in North Carolina, or omitting the surname of women who share it with their partner, removes much of the bias.

Fifth, in a nation-wide replication of a published study of the gender gap in turnout, I find that the proposed correction more than triples the estimated gap among Asian voters. Finally, I conclude by showing that the same bias arises in Lebanon, but through a different rule. Because married women in Lebanon retain their family surname but are registered to vote in their husband's locality, sex-based bias persists, but through the woman's registered locality rather than her surname.

\section{Married Surnames and Surname-Based Imputation}
\label{sec:married}

Bayesian Improved Surname Geocoding (BISG) combines the racial distribution of each surname from the Census with the racial composition of the individual's census block. Bayesian Improved First Name Surname Geocoding (BIFSG) additionally incorporates the racial distribution of the individual's first name \citep{elliott2009using, voicu2018firstname}. Both approaches assume that surname is independent of geographic location (and of first name) given race. For a woman who adopted her husband's surname, however, the surname instead reflects the race of her husband. Where she and her husband belong to the same racial group, the surname identifies her own race as expected; where the marriage is inter-racial, the surname misclassifies her.

This misclassification matters not only for the prediction of racial or ethnic group membership itself, but because of the potential correlation between the reason for misclassification (inter-racial marriage) and downstream quantities of interest. For instance, with respect to partisanship and political behavior, one might expect that the characteristics that make inter-racial marriage more likely are related to political preferences, that the characteristics that make inter-racial marriage \emph{and} surname adoption more likely are related to political preferences, and that marriage to a partner of another race shapes political preferences, through different social, economic, and political mechanisms. If women in inter-racial marriages who adopt the partner's surname differ in their partisanship or turnout, then, their misclassification is related to the outcome of interest, and the resulting errors in group-specific estimates do not average out. 

One common approach for estimating group-level downstream quantities from BISG is BIRDiE \citep{mccartan2025estimating}, which corrects the bias of estimates that weight each individual by their racial probabilities. BIRDiE requires the assumption that the outcome is independent of surname given race, geography, and observed covariates. Married surnames are, under the context just identified, precisely the situation in which a surname fails the independence assumption through a third factor, specifically the race of the spouse. I therefore expect this type of bias to persist under BIRDiE and related approaches, and to be concentrated especially among the racial groups in a target population whose women are most likely to marry men of another race and adopt their surname. 

The size of this bias can be derived simply. Consider an estimate for women of race $r$ that weights each individual by their probability of belonging to race $r$. Let $m$ denote the share of race-$r$ women who hold the surname of a husband of another race, $a_m$ and $a_o$ the average probability of race $r$ for these women and for other race-$r$ women, and $\delta$ the difference between the average outcome of other race-$r$ women and that of women who hold such a surname. Setting aside the women of other races who are assigned to race $r$ by imputation, married surnames thus shift the estimate by
\begin{equation}
\text{bias} = \frac{m(1-m)(a_o - a_m)}{m\,a_m + (1-m)\,a_o}\,\delta .
\label{eq:bias}
\end{equation}

The bias is therefore the product of three quantities: how common married surnames of another race are, how much less likely imputation is to assign the women with those surnames to their own race, and how much these women differ from other women of their race. The bias therefore vanishes when any of the three quantities is zero. Conversely, the bias is largest for groups whose women frequently marry outside the group, whose surnames are distinctive, and whose inter-married women differ politically. Because $m$ can be estimated from census data on intermarriage, the expression also allows researchers to infer the extent of the bias in settings where maiden surnames are unavailable for correction.

\section{Correcting for Married Surnames}
\label{sec:corrections}

Expression~\eqref{eq:bias} suggests two potential corrections, each of which targets a different source of the bias. The first involves recovering the surname that a woman held before marriage. Many voter files record a middle name, and in several Southern states, a considerable share of married women record their maiden surname as their middle name (Table~\ref{tab:sample}); other women hyphenate their maiden and married surnames. Where the maiden surname is recorded in a systematic way, replacing the married surname with the maiden surname restores the information that the surname holds about the woman's own race, and thus makes imputation more likely to assign women with a married surname of another race to their own race. This correction should improve both the classification of individual women and group-level downstream estimates. The correction is only available, however, where maiden surnames are recorded. 

\begin{table}[H]
\centering
\small
\begin{tabular}{lrrr}
\toprule
State & \shortstack{Women with their\\partner's surname} & \shortstack{Middle name\\is a surname (\%)} & \shortstack{Same rule\\for men (\%)} \\
\midrule
Alabama & 622,083 & 37 & 12 \\
Florida & 2,150,240 & 15 & 6 \\
Georgia & 1,191,761 & 27 & 10 \\
Louisiana & 450,173 & 42 & 6 \\
North Carolina & 1,219,096 & 42 & 12 \\
South Carolina & 674,931 & -- & -- \\
\bottomrule
\end{tabular}

\caption{Women who share their partner's surname.}
\label{tab:sample}
\end{table}

A second possible correction involves removing the married surname altogether. For women who share the surname of their partner and whose maiden surname is not recorded, the racial probabilities are then derived from geography and first name alone. Removing the married surname lowers individual accuracy, since for most married women whose husbands belong to the same racial group, the surname is informative. For group-level estimates such as those obtained with BIRDiE, however, the correction also removes the information about the partner's race that makes the surname correlated with the outcome in the first place, and thus largely restores the independence assumption on which BIRDiE relies. I therefore expect this second correction to improve estimates with BIRDiE more than it improves individual classification, and to worsen estimates most for groups whose surnames are highly informative about race, such as Black women.

A third possible correction falls in between the first two, and involves down-weighting the married surname rather than replacing or removing it. The surname can either be given less weight for every woman who shares it with her partner, or can be treated as evidence about the race of her partner, whose race relates to her own through the rates at which women of each race marry men of each race. This correction retains more of the information that the surname holds for women whose partner belongs to the same racial group, but also retains part of the information about the partner's race. I therefore expect this correction to improve individual classification at less of a cost than removing the married surname entirely, but also to improve downstream estimation with BIRDiE to a lesser extent. 

\section{Data and Design}

I use the full-count TargetSmart voter files for the six states that record self-reported race. I restrict the analysis to registered voters with self-reported race (White, Black, Hispanic, Asian, or another category), which I treat as the ground truth. The outcomes of interest are party registration (Democratic, Republican, or other) in Florida, Louisiana, and North Carolina, where such information is reported, and turnout in the 2024 general election in all six states.

To infer couples, I pair a man and a woman who are at least 20 years old and live at the same address in a household of between two and six registered voters, whose ages differ by no more than 12 years, and who each belong to exactly one such pair. The same type of bias can certainly arise in same-sex couples, in which either a woman or a man may adopt the surname of a partner of another race, though I am unable to identify such couples directly (Supplementary Section~A). I then classify each woman by her surname relative to her partner's: whether she shares her partner's surname, holds a hyphenated or compound surname that contains her partner's surname, holds a different surname entirely, or has no inferred partner. For women who share their partner's surname, I recover a maiden surname when their recorded middle name is likely a surname rather than a given name (Supplementary Note~B1). 

Because a middle name can also be a family name obtained at birth, I apply the same rule to men who share their partner's surname as a placebo. The rule flags a considerably smaller share of men than of women in every state that has usable middle names (Table~\ref{tab:sample}), which provides evidence that most of the surnames recovered for women are indeed maiden surnames. I also validate the rule directly using voter files in North Carolina, where voter files obtainable from 2005 to 2026 record the surname that each registered voter held before changing it (Supplementary Note~B2). Because South Carolina does not record usable middle names, I am only able to evaluate the corrections that do not rely on them there.

I obtain standard BISG and BIFSG probabilities from the 2020 Census surname and first-name tables \citep{census2020names} and the block-level racial composition of registered voters, under five versions of the surname used for women, which implement the corrections described in Section~\ref{sec:corrections}. The first uses every voter's current surname. The second replaces the surname of women in couples with the maiden surname where one can be recovered. The third additionally applies the middle-name rule to women without an inferred partner, who include women who are widowed, divorced, or living apart from their husband, as well as women who have never married. For a never-married woman, a surname-type middle name is most likely a family name given at birth, which typically carries information about her own race as well (Supplementary Note~B3). The fourth uses the maiden surname where one can be recovered, and otherwise omits the surname of women who share it with their partner. In this case, the racial probabilities are then inferred from first name and geography alone. I refer to this version as removing the married surname. The fifth version omits the surname of every woman who shares or hyphenates her partner's surname, and therefore does not require middle names at all. The probabilities for men remain the same in every version. I also evaluate several versions of the third correction (Supplementary Section~A).

For each version, I evaluate the probabilities in two ways. First, I compare the most probable race with self-reported race. Second, I estimate two group-level downstream quantities, the Democratic share of two-party registration and turnout, by race and sex, both by weighting each voter by their racial probabilities and with BIRDiE, conditioning on county and sex. I then compare each estimate with the same quantity computed using self-reported race. I obtain intervals by resampling counties within states.

\section{Results}

I present the results in the order of the argument. First, I quantify the extent to which different populations are misclassified by surname-based imputation, and quantify the extent to which married surnames account for this misclassification. I then demonstrate that the women misclassified on this basis differ politically from other women of their own racial group, which is identified in Section~\ref{sec:married} as the condition under which their misclassification biases downstream inference. Finally, I examine how large this bias is, how it affects group-level estimates and comparisons across age and place, and the extent to which the corrections proposed in Section~\ref{sec:corrections} rectify the issue, including in a replication of a published study.

\subsection{Who is misclassified}

\begin{figure}[H]
\centering
\includegraphics[width=\linewidth]{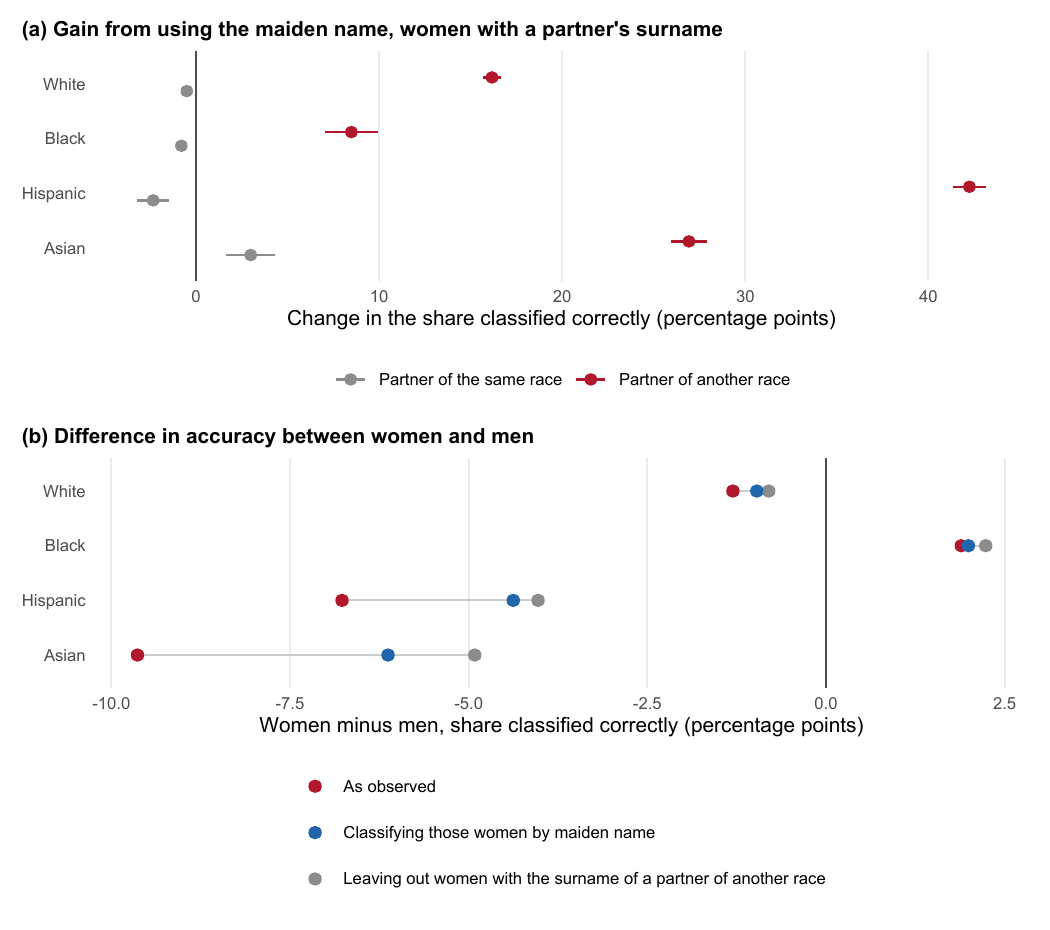}
\caption{BIFSG accuracy and married surnames. Panel (a) contains women who share their partner's surname and have a surname as their middle name. Horizontal lines are 95\% intervals. Supplementary Table~S10 gives the numbers for panel (b).}
\label{fig:swap}
\end{figure}

\begin{figure}[H]
\centering
\includegraphics[width=\linewidth]{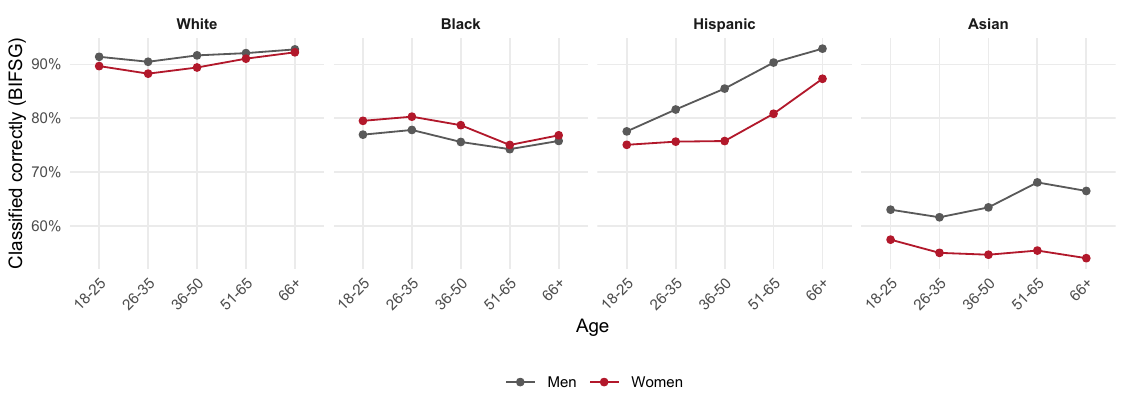}
\caption{BIFSG accuracy by age and sex, using current surnames.}
\label{fig:age}
\end{figure}

First, I quantify the extent to which different populations are misclassified by surname-based imputation. Surname-based imputation is noticeably less accurate for Asian and Hispanic women than for Asian and Hispanic men, respectively. BIFSG classifies 55.2\% of Asian women correctly, compared to 64.8\% of Asian men, and 79.0\% of Hispanic women, compared to 85.8\% of Hispanic men. The difference between men and women is smaller among White voters (90.6\% compared with 91.9\%), and reversed among Black voters (77.8\% compared with 76.0\%; Figure~\ref{fig:swap}b; Supplementary Section~A discusses calibration). Among Hispanic voters, the gap between men and women is smallest among the youngest registered voters and widens through the ages at which most women marry in the United States, ranging from 2.5 percentage points (ppts) among voters aged 18 to 25 to 9.7 ppts among those aged 36 to 50 (Figure~\ref{fig:age}). Among Asian voters, the gap is present at every age, although the gap is largest among older voters (5.6 ppts among voters aged 18 to 25, and 12.6 ppts among those aged 51 to 65). That the gap among Hispanic voters widens at the ages at which most women marry provides initial evidence that surname change at marriage accounts for at least part of the gap. 

To quantify the extent to which the married surname specifically accounts for this misclassification difference, I compare the same women under two different surnames. Among women who share their partner's surname and whose middle name is a surname, I classify each woman twice: once with her current surname, and once with her maiden surname. Women whose partner belongs to the same racial group serve as a comparison group, since for them, both surnames should be suggestive of the same racial group membership. Replacing the current surname with the maiden surname changes the share of these women classified correctly by relatively little (by between $-2.3$ and 3.0 ppts). By contrast, the replacement improves the classification of women whose partner is of another race  (Figure~\ref{fig:swap}a), often by a considerable margin. The improvement is largest for Hispanic women at 42.2 ppts, followed by Asian women (26.9 ppts). It is smaller for White women (16.2 ppts) and Black women (8.5 ppts), whose surnames hold less information about race in the first place. This provides evidence that the married surname itself, rather than other characteristics of these women, accounts for much of their misclassification. 

\begin{figure}[H]
\centering
\includegraphics[width=\linewidth]{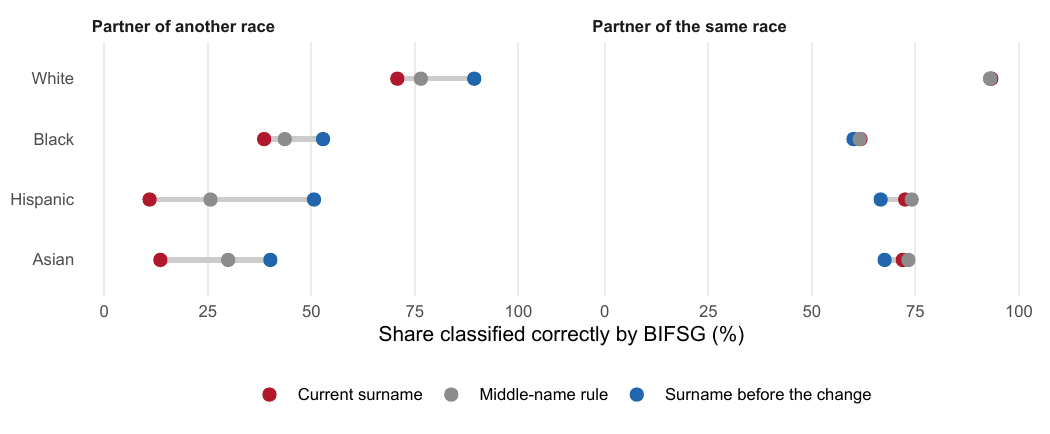}
\caption{BIFSG accuracy for women in North Carolina whose surname changed between 2005 and 2026. Panels give the race of the man who shares her current surname.}
\label{fig:nc}
\end{figure}

Because middle names record the maiden surname of only some women, I repeat this comparison using the North Carolina voter files, which record the surname that each woman held before changing it (Figure~\ref{fig:nc}). Among women whose surname changed between 2005 and 2026 and who live with a man of another race who shares her current surname, classifying each woman by the surname that she held before the change raises the share classified correctly from 10.9\% to 50.6\% among Hispanic women, and from 13.5\% to 40.1\% among Asian women. Given that accuracy is otherwise generally strong among North Carolina voters, such low accuracies among this particular subpopulation before correction indicate the severity of the bias here. Among women whose partner belongs to the same racial group, by contrast, the earlier surname does not classify them any better than does their current surname (66.6\% compared with 72.5\% among Hispanic women, and 67.6\% compared with 72.0\% among Asian women). The middle-name rule recovers only part of this improvement (25.6\% among Hispanic women and 29.9\% among Asian women), since the rule recovers only 20.0\% of these earlier surnames.

Women who hold the surname of a partner of another race thus account for a large share of the lower accuracy among Asian and Hispanic women (Figure~\ref{fig:swap}b). Because Asian women marry men of another race more often than Asian men marry women of another race \citep{pew2017intermarriage}, these women also comprise a sizable share of all Asian women (13.1\%, compared with 5.8\% of Hispanic women). Leaving them out of the analysis reduces the difference in accuracy between men and women from 9.6 to 4.9 ppts among Asian voters, and from 6.8 to 4.0 ppts among Hispanic voters. Classifying each of them by her maiden surname instead reduces the difference to 6.1 and 4.4 ppts, respectively (Supplementary Note~B4). The remainder of the difference arises among women who do not share the surname of a partner of another race, which this design does not attribute to married surnames. Omitting the surname entirely, by contrast, improves the classification of women whose partner is of another race, but worsens that of women whose partner is of the same race, for whom the surname is, as expected, informative (Supplementary Figure~S1). Thus, in the first case, the harm of dropping the surname entirely is outweighed by the harm of using the partner's surname.

\subsection{Why the error does not cancel}

\begin{figure}[H]
\centering
\includegraphics[width=\linewidth]{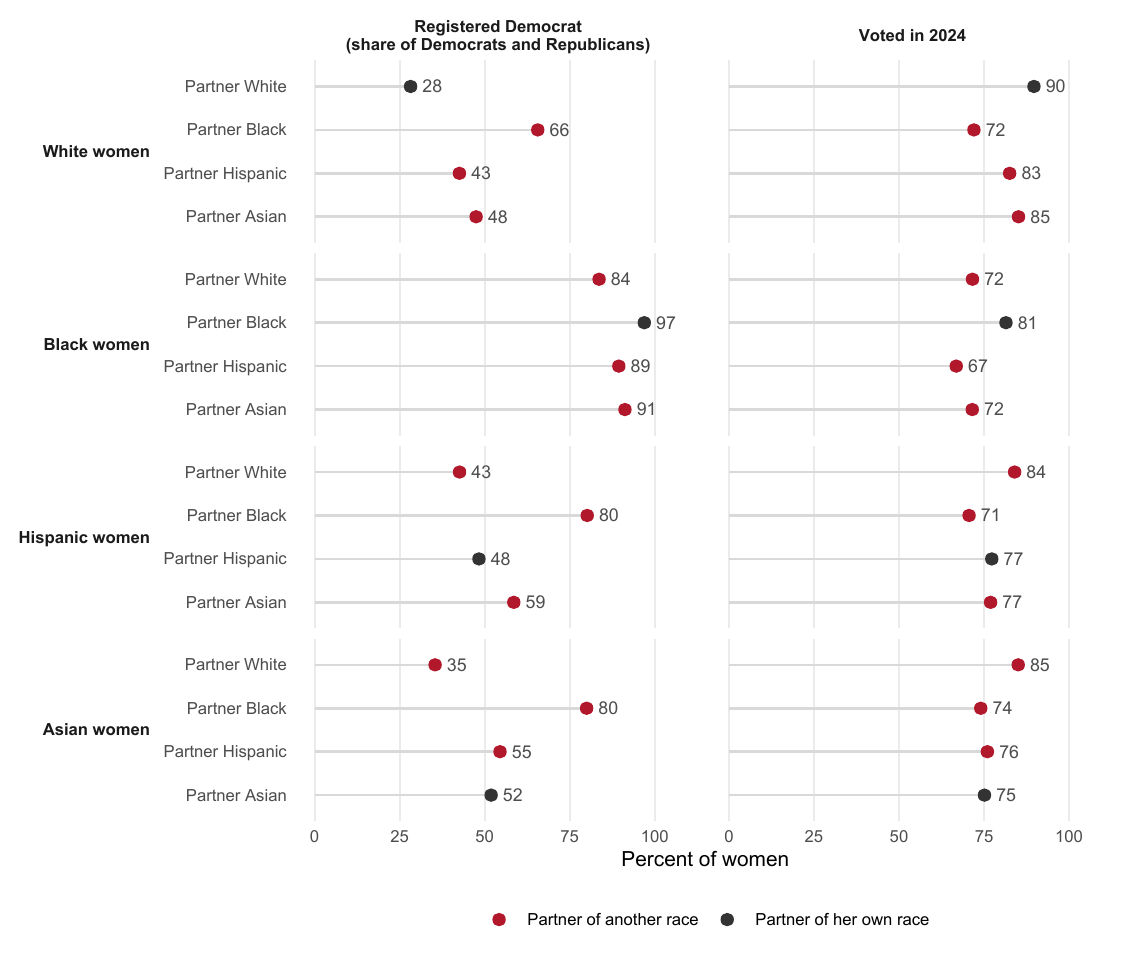}
\caption{Partisanship and turnout of women in inferred couples, by the race of the woman and of her partner. Combinations with fewer than 1,000 women are omitted.}
\label{fig:partner}
\end{figure}

As identified in Section~\ref{sec:married}, the misclassification of these women biases downstream inference only if they differ politically from other women of their own racial group. I next demonstrate that this is indeed the case for the women who are racially misclassified on this basis. Because both partners report their race, I am able to compare women of the same race by the race of their partner. Among women in inferred couples, partisanship largely follows the race of the partner (Figure~\ref{fig:partner}). Asian women whose partner is White, for instance, are considerably less likely to register as Democrats than Asian women whose partner is Asian (35.4\% of two-party registrants, compared with 51.9\%). Asian women whose partner is Black, conversely, are considerably more likely to do so (80.0\%). The same ordering holds among Hispanic women (42.6\%, 48.3\%, and 80.2\%), as well as among Black and White women. Asian and Hispanic women whose partner is White are also more likely to vote: 85.1\% of Asian women with a White partner voted in 2024, compared to 75.1\% of those with an Asian partner. Similarly, 84.0\% of Hispanic women with a White partner voted, compared to 77.3\% of those with a Hispanic partner. I am, however, unable to distinguish whether these differences reflect the characteristics that make inter-racial marriage more likely, or the influence of the partner. Nonetheless, either mechanism makes the misclassification correlated with the outcome of interest.  

Women with a White partner comprise the largest share of Asian and Hispanic women in inter-racial couples, and many of these women are assigned to the White category by imputation on the basis of their partner's surname. The estimate for all Asian women is therefore drawn disproportionately from the subset of Asian women who are more likely to register as Democrats and are less likely to vote. As a result, the estimate overstates the Democratic share of Asian women and understates their turnout. The misclassification of married women is, in this way, related to the outcome of interest, and neither weighting nor BIRDiE is able to recover the group-specific estimate. 

\subsection{How large the bias can be}

I next quantify the extent of this bias. Expression~\eqref{eq:bias} accounts for most of the bias introduced by married surnames. Among Asian women in the three states with party registration, the expression predicts that married surnames raise the weighted estimate of the Democratic share by 1.6 ppts. This is most of the bias that arises from how imputation classifies Asian women themselves, as opposed to the bias from women of other races who are assigned to the Asian category (Supplementary Note~B5). The expression does, however, understate the bias of BIRDiE; in these states, BIRDiE overstates the gender gap in partisanship among Asian voters by 4.2 ppts, which is roughly three times larger than the 1.4 ppts predicted by the expression. The expression therefore provides a conservative estimate of the extent of the bias. 

\begin{figure}[H]
\centering
\includegraphics[width=\linewidth]{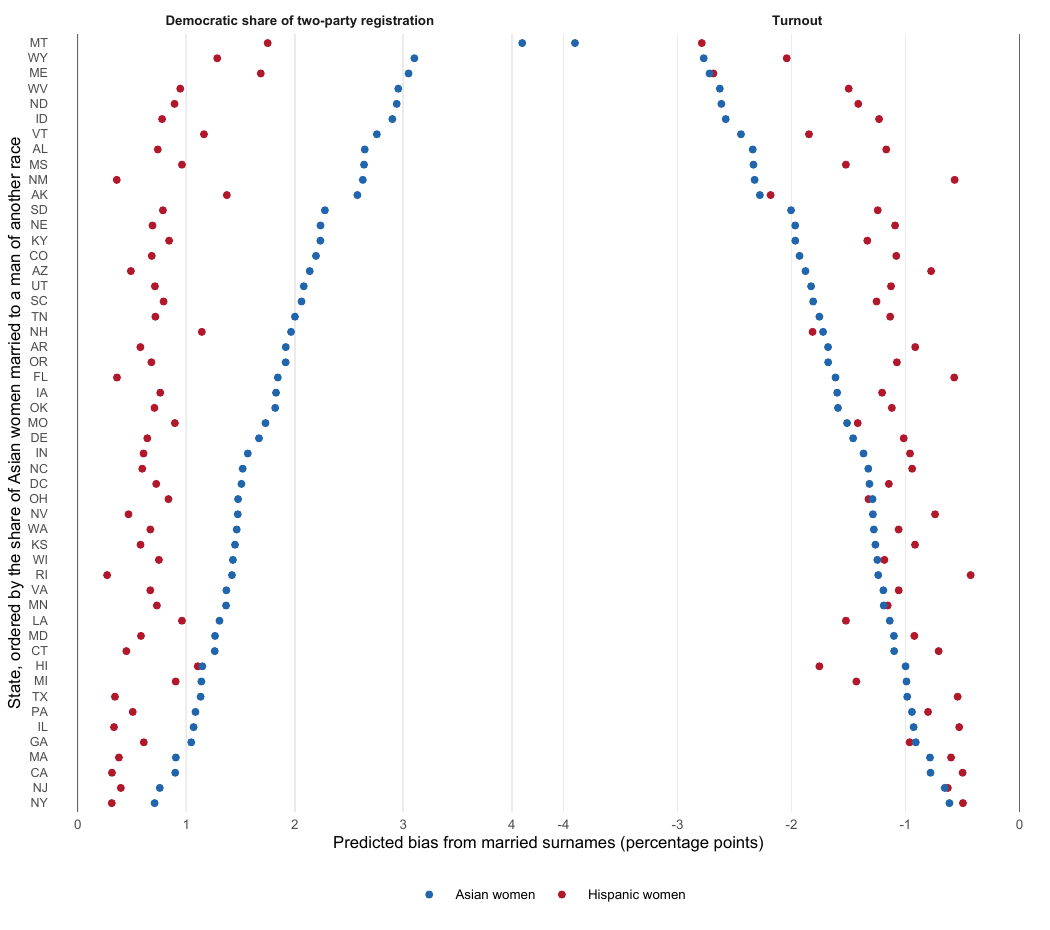}
\caption{Predicted bias in the weighted estimates for Asian and Hispanic women by state. Predictions use expression~\eqref{eq:bias} and the American Community Survey of 2022.}
\label{fig:states}
\end{figure}

Combined with the share of women of each race married to a man of another race according to the American Community Survey, the expression also yields a prediction of the bias for every state (Figure~\ref{fig:states}; Supplementary Section~A). Within the six studied states, these predictions closely track the bias observed in the voter files. Across combinations of state, age group, and race, the bias predicted from census data alone is highly correlated with the observed bias in the Democratic share (0.83) and in turnout (0.72; Supplementary Figure~S4). I find that, as expected, the bias is largest where the group is small. In states like Montana, Wyoming, Maine, and West Virginia, where a large share of Asian women are married to men of another race, the predicted bias in the Democratic share of Asian women ranges from 3.0 to 4.1 ppts, compared with 0.9 ppts in California and 0.7 ppts in New York. The predicted bias in their turnout ranges from $-2.6$ to $-3.9$ ppts, compared to $-0.8$ and $-0.6$ ppts. Surveys, moreover, contain the fewest Asian and Hispanic respondents in these settings. The bias is thus largest precisely where researchers are most dependent on voter files and individual-level inferences.

\subsection{Consequences for estimates by race and sex}

\begin{figure}[H]
\centering
\includegraphics[width=\linewidth]{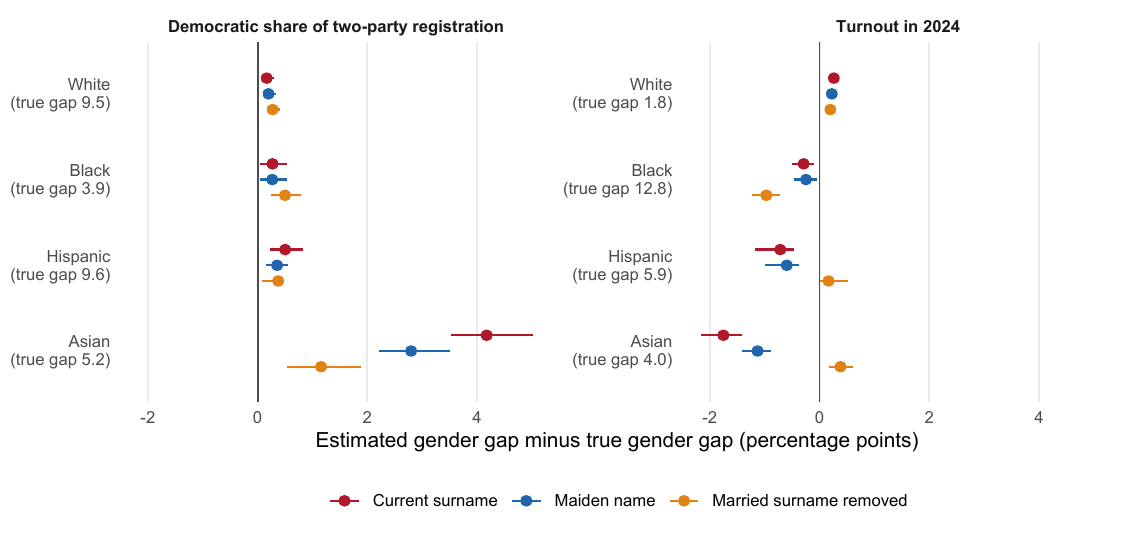}
\caption{Error in the BIRDiE estimate of the gender gap (women minus men) within each race. Horizontal lines are 95\% intervals from resampling counties.}
\label{fig:gaps}
\end{figure}

I find that married surnames bias BIRDiE estimates for Asian women and, in turn, the gender gap among Asian registered voters (Figure~\ref{fig:gaps}). Using current surnames, BIRDiE overstates the Democratic share of Asian women by 7.5 ppts, more than twice the 3.4 ppts by which it overstates that of Asian men (Supplementary Table~S7), and understates the turnout of both Asian and Hispanic women (by 2.1 and 1.2 ppts). As a result, BIRDiE nearly doubles the gender gap in partisanship among Asian registered voters (9.4 ppts, compared with 5.2 ppts computed from self-reported race), and understates the gender gap in turnout among Asian registered voters (2.3 compared with 4.0 ppts) and Hispanic registered voters (5.2 compared with 5.9 ppts). This type of bias is not driven by any single state; BIRDiE overstates the gender gap in partisanship among Asian registered voters in each of the three states with party registration, and understates the gender gap in turnout among Asian registered voters in each of the six states (Supplementary Table~S1).

Each of the proposed corrections redresses the issue to some extent, although, as anticipated in Section~\ref{sec:corrections}, the corrections that most improve individual classification differ from those that most improve group-level downstream estimates. Recovering maiden surnames reduces the overstatement of the gender gap in partisanship among Asian registered voters from 4.2 to 2.8 ppts. Additionally applying the middle-name rule to women without a partner reduces the overstatement further, to 2.1 ppts. Omitting the surname of women who share it with a partner, where no maiden surname is recovered, reduces the overstatement to 1.2 ppts, and brings the gender gap in turnout among Asian and Hispanic registered voters closer to the truth (4.4 compared with 4.0 ppts among Asian voters, and 6.1 compared with 5.9 ppts among Hispanic voters). However, this correction worsens the estimated gender gap in turnout among Black voters, for whom the surname is informative about race (an error of $-1.0$ ppts, compared with $-0.3$ ppts using current surnames).

The third correction, down-weighting the married surname, falls between the first two approaches, as expected. Treating the surname as evidence about the race of the partner improves individual classification only marginally, and does not improve estimates with BIRDiE (Supplementary Section~A and Table~S13). Because the misclassified women are concentrated at the ages at which women marry, and in the places where inter-racial couples live, BIRDiE also understates how much the gender gap in partisanship among Asian voters narrows with age, and how much the partisanship of Asian women differs between predominantly White and other neighborhoods (Supplementary Section~A).

\subsection{An application: gender gaps in turnout across congressional districts}

To quantify the extent to which this type of bias carries over to published analyses, I replicate the district-level models of  \citet{StaufferFraga2022}, who study the gender gap in turnout within racial groups across congressional districts using the individual-level race imputed by Catalist (a data vendor) on its voter file. Because their quantity of interest is precisely a gender gap within race groups, their analysis is the type most exposed to bias from married surnames. I use the nationwide TargetSmart voter files from the general elections of 2012, 2014, 2016, and 2018, covering four of the six elections that they study \citep{targetsmart2012_2018}. Because each file is constructed shortly after the respective election, turnout is measured among the registered voters of that year. For every district and year, I compute the gender gap in turnout within each race, and regress the gap on: (1) the group's share of the district's registered voters, (2) district competitiveness, (3) the Democratic share of the House vote, (4) whether a woman was one of the two leading candidates, and (5) indicators for swing states and election years included in the original models. Standard errors are clustered by district (Supplementary Section~A). I estimate each gap with BIRDiE, conditioning on district and sex, under the three versions of the surname used for women. In the states that record self-reported race, I also compute each gap from self-reported race.

\begin{figure}[H]
\centering
\includegraphics[width=\linewidth]{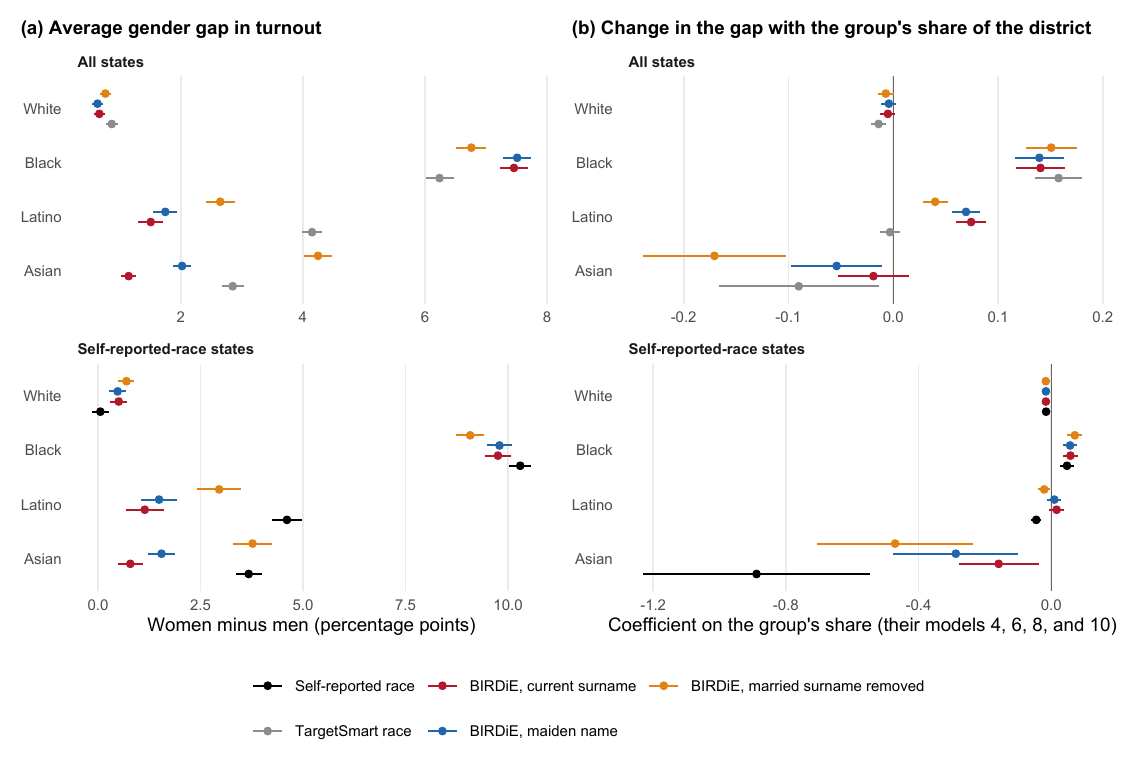}
\caption{Gender gap in turnout within each race across congressional districts, 2012--2018 (1,740 district-years). BIRDiE uses BIFSG probabilities. ``TargetSmart race'' is the race that the vendor imputes. In the bottom row, it equals self-reported race. Horizontal lines are 95\% intervals. In panel (b), they are clustered by district.}
\label{fig:sf}
\end{figure}

Using current surnames, the estimated gender gaps among Asian and Latino registered voters (1.1 and 1.5 ppts) are nearly as small as the gap among White registered voters (0.7 ppts; Figure~\ref{fig:sf}a). Removing the married surname, however, more than triples the estimated Asian gap, to 4.2 ppts, and nearly doubles the Latino gap, to 2.6 ppts. In the states that record self-reported race, the Asian gap estimated in this way is also close to the gap computed from self-reported race (3.8 compared with 3.7 ppts). Recovering maiden surnames moves the estimate in the same direction, but to a smaller extent, since middle names only correspond to maiden surnames in some states (Supplementary Note~B6). The conclusion drawn by \citet{StaufferFraga2022} that the gender gap in turnout is smallest among White voters is therefore, if anything, strengthened once married surnames are taken into account.

Married surnames also mischaracterize the extent to which the gender gaps among Asian and Latino voters vary with the composition of the district, while the coefficients for Black and White voters are nearly identical under every version (Figure~\ref{fig:sf}b and Supplementary Section~A).

\section{Generalizing Abroad: Religious Sect in Lebanon}

I conclude by demonstrating that the same type of bias arises in Lebanon, but through geography rather than surname. In Lebanon, upon marriage, a woman's place of civil registration is automatically transferred to her husband's registered locality of origin, and she votes on the list of her husband's register \citep{Chambers2009ifes}. The voter registry, however, also records her own religious sect, and she retains her own family surname. The surname of a woman listed under another sect is therefore far more typical of her own sect than of her husband's. Moreover, because the registry records both the sect of the voter and the sect of the list on which she is registered, I am able to observe directly which women have married across sects, without needing to infer couples. I therefore expect imputation to misclassify the sect of women who marry across sects through the geographic prior rather than the name, and expect a prediction from the name alone to classify these women more accurately than a prediction that also relies on locality. 

\begin{figure}[H]
\centering
\includegraphics[width=\linewidth]{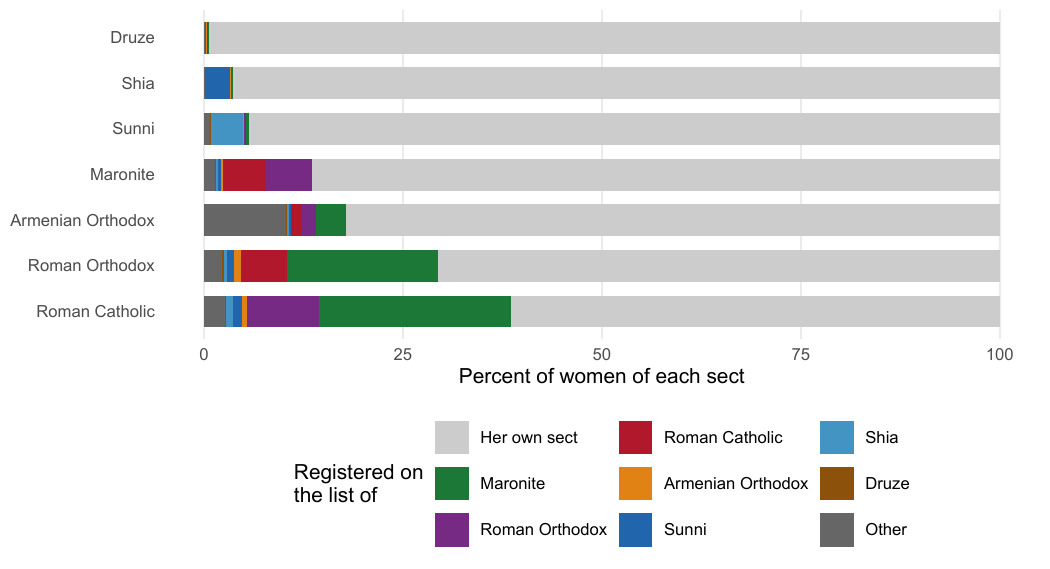}
\caption{Lebanese women of each sect by the sect of the list on which they are registered, 2018.}
\label{fig:lebflows}
\end{figure}

I use the full-count Lebanese voter file of 3.7 million registered voters \citep{dasanaike2026zeroshot}, and the sect predictions from a name-list method ($\ell$BISG) generalizing BISG outside of the United States that combines names with the sect composition of the voter's locality \citep{chasalow2026lbisg}. A considerably larger share of women than men is registered on the list of another sect (10.1\% compared with 1.5\% among the seven largest sects; Supplementary Table~S2), and among women, this share rises through the ages at which most women in Lebanon marry. Marriage across sects is concentrated among the Christian sects, particularly into the largest such sect, the Maronites (Figure~\ref{fig:lebflows}).

Women registered on the list of another sect are classified correctly far less often by $\ell$BISG than other women, at a rate of 29.4\% compared to 88.6\% (Table~\ref{tab:lebanon}). Consistent with this expectation, the source of this error is the locality of their husband's register. A prediction based on locality alone classifies 16.8\% of these women correctly; a prediction based on name alone, by contrast, classifies 52.0\% of them correctly, more than the prediction that combines name and locality. 

In Lebanon, therefore, the appropriate correction is the reverse of that in the United States, because a married woman retains her surname but is assigned the sectarian composition of her husband's locality. The registry, however, does not record marital status. Dropping the geographic prior for women in general would thus lower accuracy for the large majority of women who are registered under their own sect. This also suggests that approaches drawing primarily on names, such as zero-shot inference with large language models \citep{dasanaike2026zeroshot}, are less susceptible to this source of error than approaches that incorporate geographic information, since a married woman in Lebanon retains her own name, although such methods perform worse for women registered under their own sect, for whom locality is informative (68.8\% compared with 88.6\%). The registration of married women also affects quantities that do not involve imputation at all, such as counts of each locality's voters by the sect of their list, which misestimate the composition of mixed Christian localities (Supplementary Section~A).

\begin{table}[H]
\centering
\small
\resizebox{\linewidth}{!}{\begin{tabular}{lrrrr}
\toprule
 & \multicolumn{3}{c}{Classified correctly (\%)} & \\
\cmidrule(lr){2-4}
Registered voters & Name and locality & Name only & Locality only & $N$ (thousands) \\
\midrule
Men & 86.6 & 69.8 & 84.9 & 1,752 \\
Women & 82.6 & 67.1 & 80.1 & 1,785 \\
Women on their own sect's list & 88.6 & 68.8 & 87.2 & 1,605 \\
Women on another sect's list & 29.4 & 52.0 & 16.8 & 180 \\
\bottomrule
\end{tabular}
}
\par\medskip
\begin{minipage}{0.95\linewidth}\footnotesize ``Name and locality'' uses the predictions of \citet{chasalow2026lbisg}. ``Name only'' uses the sect distribution of each surname among men. ``Locality only'' uses the sect distribution of each locality. Both leave out the voter.\end{minipage}
\caption{Share classified correctly in the Lebanese voter registry, by sex and by whether women are registered on their own sect's list.}
\label{tab:lebanon}
\end{table}

\section{Discussion}

The adoption of surnames often follows rules of and norms around marriage rather than ethno-racial inheritance alone. Where these rules and norms differ by gender, as when wives adopt their husband's surname in the United States (or are geographically registered with their husband's family in Lebanon), probabilistic inferences of group membership misclassify women who marry across the target groups of interest. This misclassification matters not only for the prediction of group membership, but because the question of whom people marry is often endogenous to their political preferences. The resulting bias is thus correlated with downstream quantities of interest, and biases downstream inference even under methods designed to correct the bias of imputed race, such as BIRDiE. For most voters, this bias is small. For Asian and Hispanic women in the United States, however, the bias is considerable, and for the gender gaps among Asian voters in partisanship and turnout, the bias is large enough to alter substantive conclusions.

The concentration of this bias among Asian and Hispanic women (in the United States) is not incidental to its importance. These are the groups whose women most often marry outside their group, and intermarriage is more common still among younger and US-born women \citep{pew2017intermarriage}. The bias is therefore likely to grow as these cohorts comprise ever-larger shares of the electorate. The bias is also largest where these groups are small, which is the setting where surveys contain the fewest respondents and researchers are the most dependent on voter files for drawing substantive conclusions. Finally, these are the very groups whose gender gaps are studied by a growing literature using imputed race. 

I propose three corrections, which rectify the issue to different extents and are available in different data sources. Where middle names record maiden surnames, recovering them improves both the classification of individual women and group-level downstream estimates by race and sex, at little cost to women whose surname correctly identifies their race. For aggregate estimates by race and sex estimated with BIRDiE, omitting the surname of women who share it with a partner removes even more of the bias. This correction, however, lowers individual-level accuracy, and worsens some estimates for groups whose surnames are highly informative (such as the turnout rates of Black women). Down-weighting the married surname, by contrast, improves individual classification without this cost, but removes less of the bias in estimates with BIRDiE.

Where none of these corrections is feasible, expression~\eqref{eq:bias}, combined with census data on intermarriage, allows researchers to infer the likely extent of the bias. Supplementary Table~S12 also reports, for women and men of each self-reported race, the share whom imputation assigns to each race, and the replication materials report the same by state and age. I therefore recommend that researchers who report estimates by race and sex from imputed race also report the extent to which their estimates change when the surnames of married women are replaced or omitted. The same bias applies to recent extensions of BISG that predict race from name embeddings or from group-specific name lists \citep{dasanaike2026embedding, chasalow2026lbisg}, since these approaches also use the current surname as input; it also extends to approaches that use names at all, such as the zero-shot LLM-based approach of \citet{dasanaike2026zeroshot}, wherever, as in the United States, a married woman's name is her husband's surname. More generally, I expect that any attribute transmitted from one spouse to the other, such as national origin or religion inferred from names \citep{ChaturvediChaturvedi2023}, will be measured with a sex-based bias that is related to the downstream quantities of interest.

\section*{AI Disclosure}
Analysis code was prepared with the assistance of Claude (Anthropic; Claude Opus 5.5), accessed through Claude Code in September and October 2026. Opus 5.5 was also used for proofreading and light copyediting, for shortening the manuscript to meet journal limits, and for preparation of the supplementary materials. The author reviewed all code, results, and text.

\section*{Data Availability Statement}
The individual-level voter files are licensed from TargetSmart and cannot be redistributed. The Census name tables are public \citep{census2020names}, and the Lebanese voter file is described in \citet{dasanaike2026zeroshot}.

\section*{Competing Interests}
The author declares none.

\bibliography{references}

@article{argylebarber2024misclassification,
  author  = {Argyle, Lisa P. and Barber, Michael},
  title   = {Misclassification and Bias in Predictions of Individual Ethnicity from Administrative Records},
  journal = {American Political Science Review},
  year    = {2024},
  volume  = {118},
  number  = {2},
  pages   = {1058--1066},
  doi     = {10.1017/S0003055423000229}
}

@article{ImaiKhanna2016,
  author  = {Imai, Kosuke and Khanna, Kabir},
  title   = {Improving Ecological Inference by Predicting Individual Ethnicity from Voter Registration Records},
  journal = {Political Analysis},
  year    = {2016},
  volume  = {24},
  number  = {2},
  pages   = {263--272},
  doi     = {10.1093/pan/mpw001}
}

@article{ImaiOlivellaRosenman2022sciadv,
  author  = {Imai, Kosuke and Olivella, Santiago and Rosenman, Evan T. R.},
  title   = {Addressing Census Data Problems in Race Imputation via Fully {Bayesian} Improved Surname Geocoding and Name Supplements},
  journal = {Science Advances},
  year    = {2022},
  volume  = {8},
  number  = {49},
  pages   = {eadc9824},
  doi     = {10.1126/sciadv.adc9824}
}

@article{mccartan2025estimating,
  author  = {McCartan, Cory and Fisher, Robin and Goldin, Jacob and Ho, Daniel E. and Imai, Kosuke},
  title   = {Estimating Racial Disparities When Race Is Not Observed},
  journal = {Journal of the American Statistical Association},
  year    = {2025},
  volume  = {120},
  number  = {552},
  pages   = {2140--2153},
  doi     = {10.1080/01621459.2025.2526695}
}

@article{voicu2018firstname,
  author  = {Voicu, Ioan},
  title   = {Using First Name Information to Improve Race and Ethnicity Classification},
  journal = {Statistics and Public Policy},
  year    = {2018},
  volume  = {5},
  number  = {1},
  pages   = {1--13},
  doi     = {10.1080/2330443X.2018.1427012}
}

@article{elliott2009using,
  author  = {Elliott, Marc N. and Morrison, Peter A. and Fremont, Allen and McCaffrey, Daniel F. and Pantoja, Philip and Lurie, Nicole},
  title   = {Using the {Census Bureau}'s Surname List to Improve Estimates of Race/Ethnicity and Associated Disparities},
  journal = {Health Services and Outcomes Research Methodology},
  year    = {2009},
  volume  = {9},
  number  = {2},
  pages   = {69--83},
  doi     = {10.1007/s10742-009-0047-1}
}

@techreport{WordPerkins1996,
  author      = {Word, David L. and Perkins, R. Colby, Jr.},
  title       = {Building a {Spanish} Surname List for the 1990's: A New Approach to an Old Problem},
  institution = {U.S. Census Bureau, Population Division},
  type        = {Technical Working Paper},
  number      = {13},
  year        = {1996}
}

@techreport{NAACCR2011nhia,
  author      = {{NAACCR Race and Ethnicity Work Group}},
  title       = {{NAACCR} Guideline for Enhancing {Hispanic}-{Latino} Identification: Revised {NAACCR} {Hispanic}/{Latino} Identification Algorithm [{NHIA} v2.2.1]},
  institution = {North American Association of Central Cancer Registries},
  year        = {2011},
  note        = {Revised September 12, 2011}
}

@article{goldinshim2004making,
  author  = {Goldin, Claudia and Shim, Maria},
  title   = {Making a Name: Women's Surnames at Marriage and Beyond},
  journal = {Journal of Economic Perspectives},
  year    = {2004},
  volume  = {18},
  number  = {2},
  pages   = {143--160},
  doi     = {10.1257/0895330041371268}
}

@misc{pew2023namechange,
  author       = {Lin, Luona},
  title        = {About Eight-in-Ten Women in Opposite-Sex Marriages Say They Took Their Husband's Last Name},
  howpublished = {Pew Research Center, Short Read},
  year         = {2023},
  month        = sep,
  url          = {https://www.pewresearch.org/short-reads/2023/09/07/about-eight-in-ten-women-in-opposite-sex-marriages-say-they-took-their-husbands-last-name/},
  note         = {Saved as pew2023namechange.html with topline and methodology PDFs}
}

@techreport{pew2017intermarriage,
  author      = {Livingston, Gretchen and Brown, Anna},
  title       = {Intermarriage in the {U.S.} 50 Years After {Loving v. Virginia}},
  institution = {Pew Research Center},
  year        = {2017},
  month       = may
}

@article{QianLichter2011,
  author  = {Qian, Zhenchao and Lichter, Daniel T.},
  title   = {Changing Patterns of Interracial Marriage in a Multiracial Society},
  journal = {Journal of Marriage and Family},
  year    = {2011},
  volume  = {73},
  number  = {5},
  pages   = {1065--1084},
  doi     = {10.1111/j.1741-3737.2011.00866.x}
}

@article{AnsolabehereHersh2013, author={Ansolabehere, Stephen and Hersh, Eitan}, title={Gender, Race, Age and Voting: A Research Note}, journal={Politics and Governance}, volume={1}, number={2}, pages={132--137}, year={2013}, doi={10.17645/pag.v1i2.97}}

@article{StaufferFraga2022, author={Stauffer, Katelyn E. and Fraga, Bernard L.}, title={Contextualizing the Gender Gap in Voter Turnout}, journal={Politics, Groups, and Identities}, volume={10}, number={2}, pages={334--341}, year={2022}, doi={10.1080/21565503.2021.1893195}}

@article{GrumbachSahnStaszak2022, author={Grumbach, Jacob M. and Sahn, Alexander and Staszak, Sarah}, title={Gender, Race, and Intersectionality in Campaign Finance}, journal={Political Behavior}, volume={44}, number={1}, pages={319--340}, year={2022}, doi={10.1007/s11109-020-09619-0}}

@article{GhitzaGelman2020, author={Ghitza, Yair and Gelman, Andrew}, title={Voter Registration Databases and {MRP}: Toward the Use of Large-Scale Databases in Public Opinion Research}, journal={Political Analysis}, volume={28}, number={4}, pages={507--531}, year={2020}, doi={10.1017/pan.2020.3}}

@article{ChaturvediChaturvedi2023, author={Chaturvedi, Rochana and Chaturvedi, Sugat}, title={It's All in the Name: A Character-Based Approach to Infer Religion}, journal={Political Analysis}, volume={32}, number={1}, pages={34--49}, year={2024}, doi={10.1017/pan.2023.6}}

@techreport{Chambers2009ifes,
  author = {Chambers, Richard},
  title = {Voter Registration in {Lebanon}: {IFES} Briefing Paper},
  institution = {International Foundation for Electoral Systems},
  year = {2009},
  month = {February}
}

@unpublished{chasalow2026lbisg,
  title={Probabilistic Race and Ethnicity Prediction Using Group-Specific Name Lists},
  author={Chasalow, Kyla and Dasanaike, Noah and Imai, Kosuke},
  year={2026},
  note={Working paper, Harvard University}
}

@misc{census2020names,
  author = {{U.S. Census Bureau}},
  title = {Names from the 2020 Census: First Names and Last Names by Race and {Hispanic} Origin},
  year = {2020},
  note = {Data files Names2020\_FirstNames\_RaceHispanic and Names2020\_LastNames\_RaceHispanic}
}

@misc{dasanaike2026zeroshot,
  author = {Dasanaike, Noah},
  title = {Pre-Trained Language Models as Zero-Shot Tools for Social Science Research},
  year = {2026},
  note = {arXiv preprint arXiv:2610.19048},
  url = {https://arxiv.org/abs/2610.19048}
}

@misc{targetsmart2012_2018,
  author = {{TargetSmart}},
  title = {{TargetSmart} National Voter File, Post-Election Snapshots, 2012, 2014, 2016, and 2018},
  year = {2018},
  note = {Proprietary data, licensed}
}

@unpublished{dasanaike2026embedding,
  title={Using Embedding Models to Improve Probabilistic Race Prediction},
  author={Dasanaike, Noah and Imai, Kosuke},
  year={2026},
  note={Working paper, Harvard University}
}

\end{document}